\documentclass[11pt,a4paper]{article}
\pdfoutput=1
\usepackage[nohyperref]{naaclhlt2019}
\usepackage{times}
\usepackage{latexsym}
\usepackage{url}
\usepackage{amsmath}
\usepackage{graphicx}
\usepackage{CJKutf8}
\usepackage{tablefootnote}
\usepackage[ruled,linesnumbered]{algorithm2e}
\usepackage{multirow}
\usepackage{color}
 
\DeclareMathOperator*{\argmax}{arg\,max}

\newcommand{\system}{UHop\xspace}

\aclfinalcopy 
\def\aclpaperid{846} 

\title{UHop: An Unrestricted-Hop Relation Extraction Framework \\
for Knowledge-Based Question Answering}

\author{Zi-Yuan Chen$^1$, Chih-Hung Chang$^1$,  Yi-Pei Chen$^2$, Jijnasa Nayak$^3$, Lun-Wei Ku$^{1,4}$ \\
  Institute of Information Science, Academia Sinica${^1}$\\
  University of Massachusetts Amherst${^2}$\\ National Institute of Technology, Rourkela${^3}$ \\
  Most Joint Research Center for AI Technology and All Vista Healthcare${^4}$ \\
  {\tt \{zychen,lance5,lwku\}@iis.sinica.edu.tw}\\
  {\tt yipeichen@cs.umass.edu, jijnasa23@gmail.com}
}

\date{}
 
\begin{document}
 \begin{CJK*}{UTF8}{bsmi}

\maketitle
\begin{abstract}
  In relation extraction for knowledge-based question answering,
%
searching from one entity to another
 entity via a single relation is called ``one hop''. 
 In related work, an exhaustive search from all one-hop relations, two-hop
 relations, and so on to the max-hop relations in the knowledge graph is
 necessary but expensive. Therefore, the number of hops is generally
 restricted to two or three. In this paper, we propose \system, an unrestricted-hop
 framework which relaxes this restriction by use of a transition-based search
 framework to replace the relation-chain-based search one. We conduct 
 experiments on conventional 1- and 2-hop questions as well as lengthy
 questions, including datasets such as WebQSP, PathQuestion, and Grid World.
 Results show that the proposed framework enables the ability to halt,
 works well with state-of-the-art models,    
 achieves competitive performance without exhaustive searches, and opens the
 performance gap for long relation paths.
\end{abstract}

\section{Introduction}

A knowledge graph (KG) is a powerful graph structure that encodes knowledge to
save and organize it, and to provide users with direct access to this knowledge via
various applications, one of which is question answering, or knowledge-based
question answering (KBQA). 
In the knowledge graph, beliefs are commonly represented by triples showing
relations between two entities, such as LocatedIn(NewOrleans, Louisiana), where
the two entities are nodes and their relation is the edge connecting them in the
knowledge graph. Given a natural language question, a KBQA system returns
its answer if it is included in the knowledge graph; the process of
answering a question can be transformed into a traversal that starts from the
question (topic) entity and searches for the appropriate path to the answer entity.  

In the literature~\cite{Moimproved,yin2016simple,yih2015semantic} KBQA is
decomposed into topic entity linking, which determines the starting entity
corresponding to the question, and relation extraction, which finds the path to
the answer node(s).
Theoretically, relation extraction finds paths of any length,
that is, paths that contain any number of relation links, or hops (between two
nodes), as long as it reaches the answer node. 
In previous work, models consider all relation paths starting from the
topic entity ~\cite{Moimproved,yin2016simple,yih2015semantic}; we call these relation-chain-based methods. 
Two main difficulties for these methods are that processing through all
relations in a KG is not practical as the combination of these relations is
nearly infinite, and that the number of candidate paths grows exponentially
with the path length and quickly becomes intractable for large knowledge
graphs.
As a result, current relation-chain-based methods set the maximum length of
candidate paths to 1, 2 or 3. 
However, under this framework we cannot find answer entities for indirect or
complicated questions.

Most importantly, even given a larger maximum length, it is unrealistic to
expect to know in advance the maximum number of hops for real-world
applications.
Thus even with exhaustive searches, if the answer entity is still too distant or
lies outside of the search space, it is not reachable or answerable. In addition, setting a large
maximum number of hops necessitates lengthy training instances, which is
especially difficult. 

In this paper, we propose \system, an unrestricted-hop relation
extraction framework to relax restrictions on candidate path length.  
We decompose the task of relation extraction in the knowledge graph into two
subtasks: knowing where to go, and knowing when to stop (or to halt).
That is, single-hop relation extraction and termination decision.
Our contribution is threefold: 
(1) No predefined maximum hop number is required in \system, as it enables
models within the framework to halt;
(2) \system reduces the search space complexity from exponential to
polynomial while maintaining comparable results;
(3) \system facilitates the use of different models, including
state-of-the-art models. 


\section{Related Work}

State-of-the-art KBQA methods are in general 
based on either semantic parsing, or on embedding~\cite{zhou2018interpretable}.
Semantic parsing methods learn semantic parsers which parse natural language
input queries into logical forms, and then use the logical forms to query the KG for 
answers~\cite{berant2013semantic,yih2015semantic,yih2016value,krishnamurthy2017neural,iyyer2017search,peng2017maximum,socokin2018modeling}. 
These systems are effective and provide deep interpretation of the
question, but require expensive data annotation, or require training using
reinforcement learning.

Embedding-based methods first allocate candidates from the knowledge graph, represent 
these candidates as distributed embedding vectors, 
and choose or rank these vectors.
Here the candidates can be either entities or relations. Some use
embedding-based models to predict answers 
directly~\cite{dong2015question,bast2015more,hao2017end,zhou2018interpretable,lukovnikov2017neural},
whereas others focus on extracting relation paths and require further
procedures to select the answer
entity~\cite{bordes2015large,xu2016question,yin2016simple,Moimproved,Zhang2018AnAW,Yu2018KnowledgeBR,ChenRelation,shen2018chinese}.
Our work follows the latter methods in focusing on predicting relation paths,
but we seek to eliminate the need to assume in advance a maximum number of
hops.

For the solution, we turn to the field of multi-hop knowledge based reasoning.
Early methods include the Path-Ranking Algorithm and its variants.
~\cite{lao2011random,gardner2014incorporating,gardner2013improving,toutanova2015representing}
The drawback of these methods is that they use random walks independent of the type of input.
DeepPath~\cite{xiong2017deeppath} and MINERVA~\cite{das2017go} tackle this issue by framing the multi-hop
reasoning problem as a Markov decision process, efficiently searching for paths
using reinforcement learning; others propose an algorithm~\cite{yang2017differentiable} for
learning logical rules, a variational auto-encoder view of the knowledge
graph~\cite{chen2018variational,zhang2018variational}, and reward shaping technique~\cite{LinRX2018:MultiHopKG} for further improvement. 
The major difference between \system and these methods is that they do not
utilize annotated relations and hence require 
REINFORCE training~\cite{williams1992simple} for optimization.
As some datasets are already annotated with relations and paths, direct
learning using an intermediate reward is more reasonable.
Hence \system adopts a novel comparative termination decision module to
control the search process of the relation path.

The most related approach is the IRN model~\cite{zhou2018interpretable},
composed of an input module, a memory-based reasoning module, and an answer module.
At each hop, it predicts a relation path using the reasoning module, and also
optimizes it using intermediate results.
However, \system has demonstrated the ability to process large-scale knowledge
graphs in experiments conducted on Freebase~\cite{bordes2015large}.
In contrast, IRN consumes memory linearly to the size of the knowledge graph,
resulting in a limited workspace, e.g., they use a subset of Freebase in their
experiments.
Also, IRN still uses a constraint for the number of
maximum hops in the experiments, while \system needs no such limit.
Most importantly, as \system is a framework which facilitates the use of different
models,
we can expect the performance of \system to remain competitive with the
state of the art over time.

\section{UHop Relation Extraction}

\label{framework}
With \system, we aim to handle unrestricted relation hops and to be compatible with
existing relation extraction models. \system breaks down unrestricted-hop
relation extraction into two major subtasks:
single-hop relation extraction and comparative termination decision.

\begin{algorithm}
\SetAlgoLined
    Given KB, $Q$, $e$ \\
    $\mathit{stop} \leftarrow \mathit{False} $\;
    $P \leftarrow \mathit{NULL}$\;
    $R \leftarrow $ outbound relations of $e$\;
    \While{$\mathit{stop} = \mathit{False}$}{
        $\hat{r} \leftarrow $ \textbf{single hop relation extraction}\;
        $P \leftarrow P:\hat{r}$\;
        $e' \leftarrow $ traverse from $e$ through $\hat{r}$\;
        $e \leftarrow e'$\;
        $R \leftarrow $ outbound relations of $e$\;
        $\mathit{stop} \leftarrow$ \textbf{termination decision}\;
    }
    \KwResult{$P$}
    
\caption{
Unrestricted-hop relation extraction.
$e$ denotes the extracted topic entity, `$:$' is the concatenation operation, and
the termination decision returns \emph{True} if the framework decides to stop.
}
\label{algo:Uhop_algo}
\end{algorithm}

Algorithm~\ref{algo:Uhop_algo} illustrates how we perform these two tasks in
the \system framework. Given a question $Q$ and the topic entity $e$ extracted
by an existing entity linking method such as S-MART~\cite{yang2015s}, we first
query the knowledge graph for the candidate outbound relations $R$ that are connected to
$e$.
For all relations $R$, we extract single-hop relations in order to choose one
relation to transit to the next entity $e'$.
After transition ($e \leftarrow e'$), we decide whether to terminate,
that is, we determine whether the process should proceed through another
iteration to extract the next relation in the relation path. If the decision to
terminate is false, we search the KB again for outbound relations of the new
$e$, after which the search process starts again. Note that starting from the
second iteration, candidate relations are concatenated with the previously
selected relations to remember the history and consider them as a whole.
We continue this loop until the process decides to terminate. 
The termination decision thus enables \system to learn when to stop searching
for relations to extract: it determines the number of hops needed
to reach the correct target entity.
Upon termination, \system returns the extracted relation(s).

\begin{figure*}
  \includegraphics[width=\textwidth]{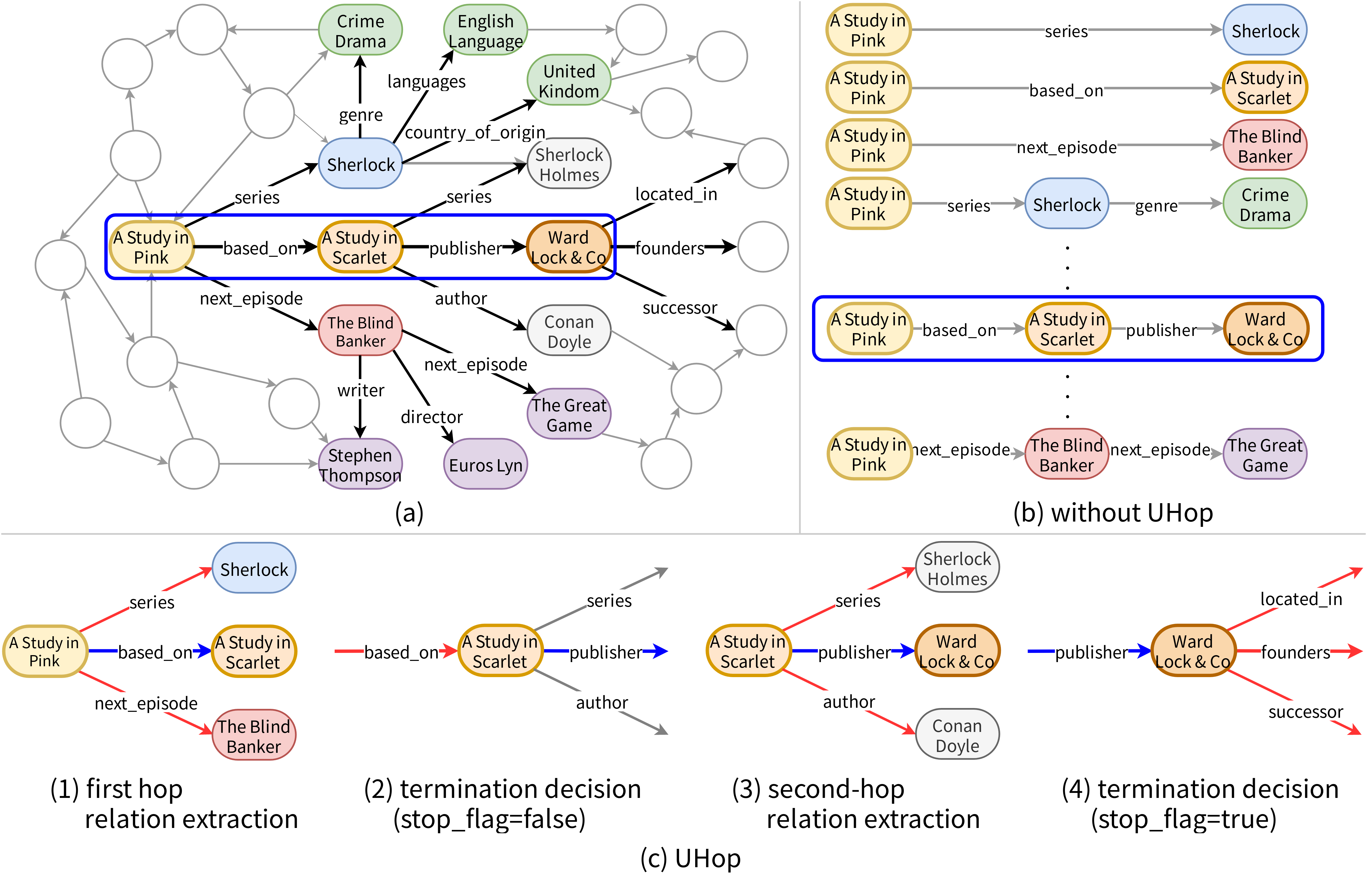}
  \caption{(a) A snippet of knowledge graph. (b) without \system, all the paths less than two hop are considered as candidates; (c) under \system, the next part of relation is extracted hop by hop (step 1 and 3), then we compare the chosen relation and its outbound relations to decide to terminate or to extract the next relation: if the extracted relation has the highest score than all the outbound relations then the process is terminated (step 4), otherwise, continued (step 2). Here we use blue arrows and red arrows to respectively represent positive/negative candidates.}
  \label{fig:kb_example}
        \vspace{-0.9pc}
\end{figure*}

In the \system framework, the model is trained to favor the correct relation
over incorrect relations.
That is, to select the correct outbound single-hop relations from current
entity $e$, the model prefers the correct $\hat{r}$ over the other relations
$R-\hat{r}$ of $e$; to terminate at the current entity $e$, the model favors
the correct relation $\hat{r}$ linked to the current entity $e$ over the
outbound $R$ relations from $e$. To continue the iteration, it proceeds likewise.
In \system, we successfully utilize this preference over relations to train the
same model to perform both single-hop relation extraction and
termination decision.
Figure ~\ref{fig:kb_example} shows the difference between previous work and our model in the scenario of multi-hop KBQA task with an simplified knowledge graph and the question ``Who published the novel adapted into A Study in Pink ?'' as example.

        \vspace{-0.5pc}

\subsection{Single Hop Relation Extraction}
\label{sec:single_hop_relation_extraction}
Single-hop relation extraction can be modeled as pairwise classification
of the set of candidate relations.
Given a question $Q$, the candidate relation set $R$, and a pairwise
classification model $F$, single-hop relation extraction is illustrated as
\begin{equation}
    r = \underset{r \in R}{\argmax}\ F(Q, r).
\end{equation}
Hinge loss, used for optimization, is defined as
\begin{equation}
    \mathcal{L}_{RE} =\frac {\sum\limits_{r\in R-\hat{r}}\max(0, -(s^{\hat{r}} - s^r) + M)}{|R-\hat{r}|},
\end{equation}
where $s^{\hat{r}}$, $s^r$ are scores of the true relation and the candidate relations
respectively. 
The margin, $M$, is an arbitrary value in the range $(0, 1]$, where the goal of the
loss function is to maximize the margin between the scores of the correct and
the incorrect predictions. 
Note that this relation extraction process and those proposed in related
work are compatible, which facilitates the installation of state-of-the-art
models in the \system framework. 

\subsection{Comparative Termination Decision}

In the \system framework, as we hope to easily replace the used model by
state-of-the-art models, we make the termination decision using the same
model for single-hop relation extraction so that no additional model is
needed.
Therefore, we propose a progressive method which treats the termination
decision as a comparison.
That is, the model stops when it cannot extract any relation better than that
from its previous hop.

What is different here is $R$, the relations to be compared against $\hat{r}$,  are the concatenation of extracted relation and all the relation starting from the new current entity $e$;  recall that we update $e \leftarrow e'$ before we step into termination decision.
If the score $s^{\hat{r}}$ is higher than all the
compared relations, the searching process terminates; otherwise, it
continues.

Given a question $Q$, an extracted relation $\hat{r}$ from the previous entity,
the candidate relation set $R$ from the new current entity $e$, and the same model $F$ as in the single hop relation extraction,
the procedure can be formulated as
\begin{equation}
    \mathit{stop} = 
\begin{cases}
    \mathit{True},& F(Q, \hat{r}) > F(Q, r) \:\: \forall r \in R\\
    \mathit{False},& F(Q, \hat{r}) < F(Q, r) \:\: \exists r \in R
\end{cases}
\end{equation}

Loss is defined depending on the flag \textit{stop}.
If the process should continue, i.e., \textit{stop} is false, loss is
defined as
\begin{equation}
    \mathcal{L}_{TD} = \max(0, -(s^{r'} - s^{\hat{r}}) + \text{margin}),
\end{equation}
where score $s^{r'}$ is the score of the question paired with the gold relation $r'$ in the next hop
and $s^{\hat{r}}$ is the score of the question paired
with the extracted relation $\hat{r}$.
In contrast, if the process should terminate, we optimize the model by
\begin{equation}
    \mathcal{L}_{TD} = \frac{\sum\limits_{r \in R}\max(0, -(s^{\hat{r}} - s^r) + M)}{|R|}. 
\end{equation}
The model thus learns to infer $s^{\hat{r}}$ is greater than $s^r$, resulting
in the termination of relation extraction.  

\subsection{Dynamic Question Representation}
While \system inferences hop by hop, it is
straightforward to enforce the focus at different aspects of the question. For this purpose, we update the question representation for each hop by defining a dynamic question representation generation function $G$.
Given the previously selected relation path $P$ and the original question $Q$, $G$
generates the new question representation as $Q^\prime = G(Q, P)$.
Our assumption is that since the current relation has been selected,
its related information in the question loses importance when
extracting the next relation. 

Inspired by both supervised
attention~\cite{mi2016supervised,liu2016neural,kamigaito2017supervised}, which
is lacking in our datasets, and the coverage loss design for
summarization~\cite{see2017get}, we de-focus the selected relation by
manipulating weights in the question representation.
We propose two ways of updating the question representation, taking into account the existence of the
attention layer in the model's architecture.
For attentive models, we directly utilize the attention weight as
part of our dynamic question representation generation function by
\begin{equation}
    G(Q, P) = W(Q - \mathit{attention}(Q, P)) + B.
\end{equation}

For non-attentive models, we apply a linear transformation
function as $G$ on the concatenation of the previously selected relation and the
question representation to yield the new representation:
\begin{equation}
    G(Q, P) = W[Q:P] + B,
\end{equation}
where $W$ and $B$ are weight matrices to be
optimized during training.

\subsection{Jointly Trained Subtasks}
In training, we jointly optimize the two subtasks of \system. For each question and its candidates, the loss function is defined as
\begin{equation}
    \mathcal{L} = \sum^H_i (\mathcal{L}^{(i)}_{RE} + \mathcal{L}^{(i)}_{TD}),
\end{equation}
where $H$ is the number of hops in the gold relation path; $\mathcal{L}^{(i)}_{RE}$ and $\mathcal{L}^{(i)}_{TD}$ are the loss of the two subtasks at the $i$-th hop respectively.

\section{Experiments} \label{experiment}
In this section, we illustrate the performance \system achieves
while reducing the search space, and its relation inference power for multi-hop
questions. Performances of the state of the art models are listed as the upper-bound. 

\subsection{Datasets} \label{dataset}
For our benchmarking evaluation materials, we selected WebQSP
(WQ)~\cite{yih2016value}, as it is used in most related work. WebQSP is the
annotated version of WebQuestions~\cite{berant2013semantic}, which contains
questions that require a 1- or 2-hop relation path to arrive at the answer
entity. More specifically, about 40\% of the questions require a 2-hop
relation to reach the answer.
This dataset is based on the Freebase knowledge graph~\cite{bordes2015large}.
For questions with multiple answers, we use each answer to construct a
question-answer pair.
Every question is annotated with its inferential relation chain (i.e., a
relation), topic entity, and answer entity. 
The statistics for these two datasets are shown in Table~\ref{tab:pq_statistic}. 

As WQ contains only questions with 1- and 2-hop answers that are still short, we also conduct
experiments for path length related analysis on the PathQuestion
dataset~\cite{zhou2018interpretable}, which includes questions requiring 3-hop
answers. 
To the best of our knowledge, this is the only available general-KB dataset
containing 3-hop questions.
PathQuestion provides two datasets: PathQuestion (PQ) and PathQuestion-Large
(PQL). 
These both contain 2-hop (PQ2/PQL2) and 3-hop (PQ3/PQL3) questions
respectively, and both use a subset of Freebase as their knowledge graph.
Note that for both PQ and PQL, questions are generated using templates,
paraphrasing, and synonyms.
PQL is more challenging than PQ because it
utilizes a larger subset of Freebase, and provides fewer training instances.
Table~\ref{tab:pq_statistic} shows statistics of these datasets.
\vspace{-0.5pc}
\begin{table}[ht]
    
    \centering
    \begin{tabular}{ |c|c|c c c|c| } 
         \hline
          & hops & Train & Valid & Test  \\
         \hline
         \multirow{2}{*}{WQ} & 1 & 2113 & - & 1,144 \\
          & 2 & 1,285 & - & 647 \\
         \hline
         PQ2 & 2 & 1,526 & 191 & 191 \\
         PQ3 & 3 & 4,158 & 520 & 520 \\
         \hline
         PQL2 & 2 & 1,275 & 159 & 160  \\
         PQL3 & 3 & 1,649 & 206 & 207 \\
         \hline
         Grid 2--4 & 2--4 & 68,046 & 9,742 & 19,298\\
         Grid 4--6 & 4--6 & 73,092 & 10,362 & 21,037 \\
         Grid 6--8 & 6--8 & 41,473 & 5,844 & 11,789 \\
         Grid 8--10 & 8--10 & 18,386 & 2,667 & 5,326 \\
         \hline
\end{tabular}
    \caption{Number of questions in experimental datasets}
    \label{tab:pq_statistic}
    \vspace{-1.0 pc}

\end{table}

The above datasets serve to show that the \system
framework yields performance competitive with state-of-the-art KBRE
models.
Further, we seek to demonstrate that \system reduces the
search space when required reasoning paths are even longer, i.e., 
longer than 3 hops, and that \system works for different kinds of relations. 
For this we use Grid World~\cite{yang2017differentiable},
a synthetic dataset with questions requiring lengthy~-- up to 10 hops~-- 
relation paths to answer. We select it to demonstrate that \system works for
long as well as task-specific relations.
In Grid World, the input is the starting node, a sequence of navigation
instructions, and a 16-by-16 fully connected grid. The model must
follow the instructions to arrive at the destination node.
Specifically, the task is to navigate to an answer cell (answer entity)
starting from a random cell (topic entity) given a sequence of instructions
(questions).
The KB consists of triples such as ((4, 1), South, (5, 1)), which indicates that the
entity (5, 1) is south of the entity (4, 1); questions are
sequences of directions such as (North, NorthEast, South).
Samples in Grid World are classified into 4 buckets~-- [2--4], [4--6], [6--8], and
[8--10]~-- according to their reasoning path length. Unlike relations included
in general knowledge bases like Freebase, relations in Grid World are the
relative directions of two nodes.

MetaQA~\cite{zhang2018variational}  and sequence QA are two other multi-hop
knowledge-based question-answering datasets which we do not use for
experiments in this paper.
MetaQA is a multi-hop dataset for end-to-end KBQA
based on a movie knowledge graph with 43k entities. 
However, it is too simple for discussions as it contains only 6 relations and on average the number of the outbound relations for each node is 3. 
The Complex Sequential QA dataset~\cite{saha2018complex} improves
on overly simplistic KBQA datasets.
Nevertheless, instead of questions requiring multi-hop relation paths, it provides
a sequence of questions, each of which requires a single-hop relation to
answer, resulting a different setting. Hence these two datasets are beyond the scope of this paper.


\subsection{Benchmark: WQ Experiments}
\subsubsection{Baseline and Settings}

We used two state of the art models, HR-BiLSTM~\cite{Moimproved} and ABWIM~\cite{Zhang2018AnAW}, as the models for use within the \system
framework. Another state of the art model, MVM~\cite{Yu2018KnowledgeBR}, is not selected here as it requires additional information:
the tail entity type. In MVM, to consider each $n$-th-hop relation, the
model searches all related $(n+1)$-th-hop relations to collect enough
information; thus further queries are necessary in MVM. 
This property of MVM causes the \system to degrade to a
relation-chain based model, which we are trying to avoid.

We report the results of these two models working within and independent of the \system
framework to evaluate whether relaxing the constraint on the number of hops has
any impact on their performance. 
For comparison, we select BiCNN as baselines and list their
results. 
As there is no pre-defined validation set in WQ, we randomly select 10\% of the
training data as the validation set.
The best parameters for different models and datasets were set empirically.

In all cases we used 300-dimensional pretrained GloVe~\cite{pennington2014glove}
word embeddings and RMSprop optimization. In ABWIM, following the setting of \cite{Zhang2018AnAW}, we
respectively chose 1, 3, 5 as kernel sizes and 150 as the number of filters for its three CNN layers.
We tune the following hyperparameters with grid search : (1) the hidden size for all LSTM ([100, 150, 256]); (2) dropout rate ([0, 0.2, 0.4]); (3) margin for Hinge loss ([0.1, 0.3, 0.5, 0.7, 1.0]); (4) learning rate ([0.01, 0.001, 0.0001]).

\begin{table}[!ht]
    \centering
    \begin{tabular}{ |c|c| } 
         \hline
         Method & Accuracy  \\
         \hline
         BiCNN~\cite{yih2015semantic}  & 77.74 \\ 
         
         HR-BiLSTM~\cite{Moimproved}  & 82.53 \\
         
			ABWIM~\cite{Zhang2018AnAW} & \phantom{ }83.26\tablefootnote{Note that the
			original paper reported 85.32, but we failed to reproduce such
			performance. Hence we report our reproduced performance which is the
			same model adapted in our proposed framework.}  \\
		\hline
		    HR-BiLSTM with UHop & 82.60  \\
         ABWIM with UHop & 82.27 \\
         
         \hline
\end{tabular}
	 \caption{Results adopting state-of-the-art models in \system framework
	 vs standalone versions}
    \label{tab:main_exp_result}
\end{table}

\subsubsection{Results and Discussion}

The experimental results are shown in Table~\ref{tab:main_exp_result}.
As expected, the performance of models within the \system framework is comparable to
those independent of it, with the additional advantage of the unrestricted number of
relation hops and a greatly reduced search space.

Table~\ref{tb:search_space} lists the average number of candidates the experimental
models consider for each question when working within and independent of \system. 
For a dataset based on a KB with an average of $n$ relations connected
to each entity, the approximate search space without \system is $n(n-1)^{(L-1)}$, where
$L$ is the predefined maximum hop number; with \system the approximate search
space is reduced to $n(L+1)$.
The specific number depends on the actual number of outbound relations connected to the
entities.
Table~\ref{tb:search_space} shows that \system reduces the search space by 30\%
for WQ, which translates to lower processing time, less memory consumption, and
sometimes slightly improved performance. 
\begin{table}[h]
    \centering
    \begin{tabular}{|c|c|c|}
    \hline
    & Train & Test\\
    \hline
    Without \system &  97.2 & 98.8 \\
    With \system &  66.7 & 65.6 \\
       
    \hline
    \end{tabular}
    \caption{Number of relation candidates in WQ}
    \label{tb:search_space}
    \vspace{-1.0pc}
\end{table}


\subsection{More Hops: PQ/PQL Experiments}

\begin{table*}[ht!]
    \centering
    \begin{tabular}{ |c|c c c|c c c| } 
         \hline
         Method & PQ2 & PQ3 & PQ+ & PQL2 & PQL3 & PQL+  \\
         \hline
         IRN~\cite{zhou2018interpretable} & 96\phantom{.00} & 87.7\phantom{0} & 53.6\phantom{0} & 72.5\phantom{0} & 71\phantom{.00} & 52.9\phantom{0} \\
         \hline
         HR-BiLSTM~\cite{Moimproved} & 100 & 99.62 & 99.72 & 97.5\phantom{0} & 87.92 & 92.92 \\
         HR-BiLSTM with UHop & 99.48 & 99.23 & 99.72 & 91.25 & 88.41 & 91.01 \\
         HR-BiLSTM with UHop + DQ & 100 & 99.62 & 99.58 & 95\phantom{.00} & 89.37 & 91.83 \\
         \hline
         ABWIM~\cite{Zhang2018AnAW} & 98.95 & 99.81 & 99.72 & 94.37 & 89.37 & 92.64 \\
         ABWIM with UHop & 97.38 & 99.62 & 99.02 & 91.25 & 88.89 & 91.01 \\
         ABWIM with UHop + DQ & 100 & 99.62 & 99.44\phantom{0} & 97.5\phantom{0} & 89.37 & 92.37\\
         \hline
\end{tabular}

	 \caption{Accuracy on PathQuestion. PQ+ is mix of PQ2 and PQ3, and PQL+
	 contains PQL2 and PQL3. We use the accuracy reported in
	 \cite{zhou2018interpretable} directly for PQ2, PQ3, PQ2L and PQ3L; for
	 PQ+ and PQL+ we use the model released with the dataset. DQ stands for
	 dynamic question representation.}
    \label{tab:exp2_result}

\vspace{-1em}
\end{table*}
\subsubsection{Baseline and Settings}
Following the original paper~\cite{zhou2018interpretable}, PQ and PQL are both
partitioned into training/validation/testing sets at a ratio of 8:1:1. 
In addition to the original PQ/PQL dataset, we merge PQ2 and PQ3, and then PQL2
and PQL3, to create the mixed datasets PQ+ and PQL+ to evaluate if the
model terminates correctly instead of always 
stopping on the majority of the training data length. 
Again we adopt HR-BiLSTM and ABWIM in this experiment.
In addition, the IRN model\footnote{https://github.com/zmtkeke/IRN. We 
consulted the authors of the repository, who stated that this version is
not the one in their paper, which they did not release publicly.} proposed
together with the PQ/PQL dataset was selected as one of the baselines for
comparison. For this dataset containing questions of long relation paths, we
also applied the dynamic question representations (DQ) in \system.

\subsubsection{Results and Discussion}

Results\footnote{Note that IRN's performance was evaluated using final answer
prediction, which is slightly different from relation path prediction.
However, finding the correct relation path should imply finding the correct
answer entity. 
} are shown in Table~\ref{tab:exp2_result}. 
Both HR-BiLSTM and ABWIM either within or independent of \system outperform IRN
and perform nearly perfectly in all datasets, which confirms that \system is
competitive even with longer relation paths. However, as shown in
Table~\ref{tab:pq_search_space}, the search space reduction for PQ and PQL is
not obvious. We find that the knowledge
graph used in PQ/PQL (a subset of Freebase) is much smaller and less
complicated than the original Freebase used in WQ, i.e., the outbound degree of nodes is relatively small.
Nevertheless, \system still performs comparably with previous work. This
indicates that it also works well in small and simple KBs.

As all PQ/PQL questions are multi-hop questions, we used dynamic
question representations to better reflect transitions in the relation
extraction process.
Table~\ref{tab:exp2_result} shows that updating the question representation
dynamically (+DQ) in each iteration benefits relation extraction in most
cases.
\begin{table}[h]
    \centering
\resizebox{\columnwidth}{!}{
    \begin{tabular}{|c|c|c|c|c|c|c|c|}
    \hline
    \multicolumn{2}{|c|}{} & \multicolumn{3}{c|}{2-hop} & \multicolumn{3}{c|}{3-hop} \\
    \cline{3-8}
    \multicolumn{2}{|c|}{} & Train & Valid & Test & Train & Valid & Test \\
    \hline
    \multirow{2}{*}{PQ}
    & W/o & 3.53 & 3.65 & 3.90 & 9.77 & 9.63 & 9.61\\
    & With & 3.68 & 3.81 & 3.85 & 9.14 & 9.27 & 9.48\\
    \hline
    \multirow{2}{*}{PQL}
    & W/o  & 3.71 & 2.64 & 3.91 & 24.52 & 12.96 & 8.45\\
    & With & 3.94 & 3.23 & 4.29 & 10.11 & 9.28 & 6.98\\
    \hline
    \end{tabular}
}
    \caption{Candidates of PQ/PQL within and independent of the \system framework}
    \label{tab:pq_search_space}
            \vspace{-1.0pc}

\end{table}


\subsection{Very Long Paths: Grid World}
\subsubsection{Baseline and Settings}

In the Grid World experiments, we used MINERVA~\cite{das2017go} and Neural
LP~\cite{yang2017differentiable} as baselines.
As understanding questions is not an issue here, we randomly initialized the word embeddings
and optimized them during the training process. 
We set the learning rate to 0.001, the hidden size to 256, the embedding size
to 300, and optimized the model using the RMSprop~\cite{hinton2012neural}
Algorithm. In this experiment, the search space has gone too large to afford for HR-BiLSTM and ABWIM without the assistance of \system.

\begin{figure}[ht]
\centering
\includegraphics[width=\columnwidth]{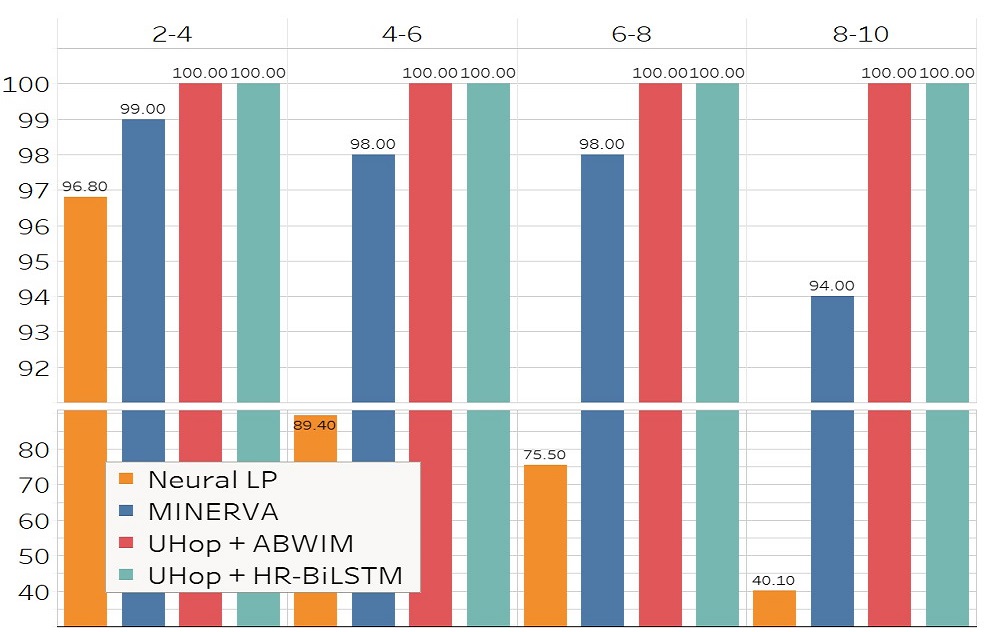}
\caption{Grid World results of state-of-the-art knowledge-based relation
extraction models} \label{fig:grid_world_bar}
        \vspace{-0.5pc}

\end{figure}
 \vspace{-0.5pc}
\subsubsection{Results and Discussion}
The results in Figure~\ref{fig:grid_world_bar} show that together with the
relation extraction model, \system perfectly solves this problem. In the
first place, compared to Neural LP and MINERVA, \system benefits from the more
powerful natural language understanding models~-- HR\_BiLSTM and ABWIM~--
equipped with sophisticated LSTM models, whereas Neural LP and MINERVA only use
multi-layer neural networks as the policy network. This demonstrates \system's
merit of facilitating the use of novel models. In the second place,
Figure ~\ref{fig:grid_world_bar} shows that error propagation 
leading to poor performance for long-path questions in Neural LP and MINERVA is
mitigated by the relation inference power of \system: it performs well for all
four buckets of questions.
Also, as Grid World includes paths of up to 10 hops, conducting experiments
purely by relation-chain based models themselves like HR-BiLSTM or ABWIM
independent of \system is not feasible: the number of candidate relations in
the exhaustive search space grows exponentially.
In Grid World, there are 8 directions (relations), and models are allowed to go back and forth. Hence given the path length $k$, the approximate search space for the models working independently is $8^k$, while for models working within \system is $8\times k$.  We observe that without \system, the required search space would preclude
experiments even on the set containing the shortest paths (Grid World [2--4]),
much less the longer ones.

\section{Further Discussion}
\subsection{Dataset Characteristics}

In this section we further compare the experimental multi-hop KBQA datasets
WQ, PQ, and Grid World.
Grid World contains questions that require the longest reasoning
paths.
However, they are synthetic, the relations are simply direction tokens, and the
questions are just sequences of direction instructions.
Therefore in this paper, it is only used to test the model's ability of making long sequential
decisions instead of understanding questions. From experiments we have seen that delicate models like HR-BiLSTM and ABWIM cannot work on it without \system, and other models such as Neural LP and MINERVA perform worse as they are rewarded only by question.

On the other hand, in WQ, questions are written in natural language and can be answered by
1-hop or 2-hop reasoning.
However, for real-world questions, 2-hop reasoning is still overly simplistic. 
For example, although WQ questions such as ``What is the name of Justin
Bieber's brother?'' are challenging for models,
humans can easily answer these with a simple Internet search.

\begin{figure}[h]
\centering
\includegraphics[width=\columnwidth]{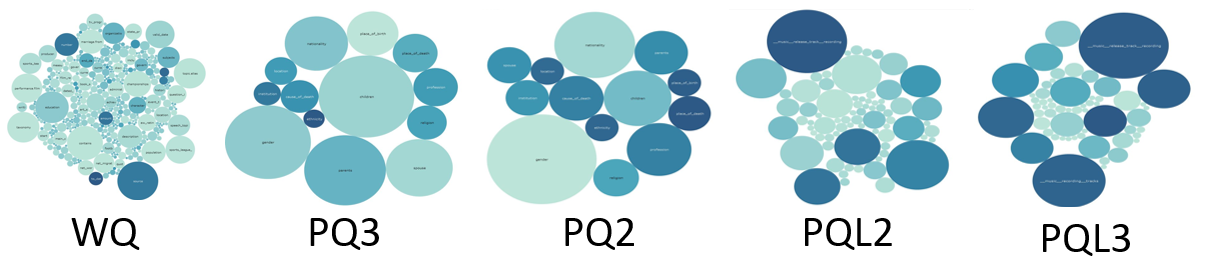}
 \vspace{-1.0pc}
\caption{A visualization of the KB relations that cover the dataset. The bubble's size is proportional to the relation's frequency.}
\label{fig:relation_distribution}
 \vspace{-0.5pc}
\end{figure}

Noting this problem, the authors of IRN~\cite{zhou2018interpretable} propose PQ 
and PQL, for which questions require at least 2-hop at most 3-hop relation paths. 
However, PQ/PQL also has its limitations.
First, the KB used in PQ/PQL is smaller than that in WQ, and its
relations are repetitive and show little variety.
Figure~\ref{fig:relation_distribution} illustrates the relation distributions.
Second, PQ/PQL questions are generated by extracting relation paths and
filling templates, which can lead to questions with obvious, learnable patterns.
This can be observed by comparing results in
Tables~\ref{tab:main_exp_result} and \ref{tab:exp2_result}. However, 
repeated relations could also help the model to learn better dynamic
question representations with respect to these relations.
Table~\ref{tab:exp2_result} shows that updating question representations
dynamically (DQ) does improve PQ/PQL performance.

\vspace{-0.5pc}
\subsection{Trained on 3-hop, Tested on 2-hop}
To evaluate if the model halts in the search process, we conducted an
experiment using PQL3 as the training/validation set and PQL2 as the
testing set. The results are shown in Table~\ref{tab:exp4}. Within the \system 
framework, both models outperform their original version by more than 7\%.
However, with zero 2-hop samples, it still overfits on the 3-hop length in training data, resulting in
accuracies lower than 50\%.
\begin{table}[ht]
    \centering
    \begin{tabular}{ |c|c|c| } 
         \hline
         & HR-BiLSTM & ABWIM \\
         \hline
         Without \system & 32.18 & 32.94 \\
         With \system & 39.65 & 49.94 \\
         \hline
\end{tabular}
	 \caption{Accuracies of models trained on PQL3 and tested on PQL2. The maximum length of relation paths for
	 models without \system is set to 3.}
    \label{tab:exp4}
    \vspace{-1em}
\end{table}

\subsection{Error Analysis}
\begin{table}[ht]
\centering
\resizebox{\columnwidth}{!}{
\begin{tabular}{|c|c|c|c|c|c|c|c|c|}
\hline
\multicolumn{3}{|c|}{\multirow{2}{*}{Dataset/model}} & \multicolumn{2}{c|}{1-hop} & \multicolumn{2}{c|}{2-hop} & \multicolumn{2}{c|}{3-hop} \\ \cline{4-9} 
\multicolumn{3}{|c|}{} & RE & TD & RE & TD & RE & TD \\ \hline
\multirow{4}{*}{WQ} & \multirow{2}{*}{1-hop} & H & 17.46 & 0 & - & - & - & - \\ \cline{3-9} 
 &  & A & 20.95 & 0.1 & - & - & - & - \\ \cline{2-9} 
 & \multirow{2}{*}{2-hop} & H & 16.15 & 1.03 & 1.2 & 0 & - & - \\ \cline{3-9} 
 &  & A & 18.38 & 0.86 & 2.06 & 0.17 & - & - \\ \hline
\multicolumn{2}{|c|}{\multirow{4}{*}{PQ2}} & H & 0.52 & 0 & 0 & 0 & - & - \\ \cline{3-9} 
\multicolumn{2}{|c|}{} & H* & 0 & 0 & 0 & 0 & - & - \\ \cline{3-9} 
\multicolumn{2}{|c|}{} & A & 2.09 & 0 & 0.52 & 0 & - & - \\ \cline{3-9}
\multicolumn{2}{|c|}{} & A* & 0 & 0 & 0 & 0 & - & - \\ \hline
\multirow{8}{*}{PQ+} & \multirow{4}{*}{2-hop} & H & 0 & 0 & 0 & 0 & - & - \\ \cline{3-9} 
 &  & H* & 0 & 0 & 0 & 0 & - & - \\ \cline{3-9} 
 &  & A & 0 & 0 & 0.52 & 0 & - & - \\ \cline{3-9} 
 &  & A* & 0 & 0 & 0 & 0 & - & - \\ \cline{2-9} 
 & \multirow{4}{*}{3-hop} & H & 0 & 0 & 0 & 0 & 0.38 & 0 \\ \cline{3-9} 
 &  & H* & 0 & 0 & 0 & 0 & 0.58 & 0 \\ \cline{3-9} 
 &  & A & 0 & 0 & 0 & 0 & 1.15 & 0 \\ \cline{3-9}
 &  & A* & 0 & 0 & 0 & 0 & 0.77 & 0 \\ \hline
\multicolumn{2}{|c|}{\multirow{4}{*}{PQ3}} & H & 0 & 0 & 0.19 & 0 & 0.58 & 0 \\ \cline{3-9} 
\multicolumn{2}{|c|}{} & H* & 0 & 0 & 0 & 0 & 0.38 & 0 \\ \cline{3-9} 
\multicolumn{2}{|c|}{} & A & 0 & 0 & 0 & 0 & 0.38 & 0 \\ \cline{3-9}
\multicolumn{2}{|c|}{} & A* & 0 & 0 & 0 & 0 & 0.38 & 0 \\ \hline
\multicolumn{2}{|c|}{\multirow{4}{*}{PQL2}} & H & 5.62 & 0 & 3.12 & 0 & - & - \\ \cline{3-9} 
\multicolumn{2}{|c|}{} & H* & 2.5 & 0 & 2.5 & 0 & - & - \\ \cline{3-9} 
\multicolumn{2}{|c|}{} & A & 5.0 & 0 & 3.75 & 0 & - & - \\ \cline{3-9}
\multicolumn{2}{|c|}{} & A* & 0 & 0 & 2.5 & 0 & - & - \\ \hline
\multirow{8}{*}{PQL+} & \multirow{4}{*}{2-hop} & H & 0 & 0 & 3.12 & 0 & - & - \\ \cline{3-9} 
 &  & H* & 0 & 0 & 3.75 & 0 & - & - \\ \cline{3-9} 
 &  & A & 0 & 0 & 3.75 & 0 & - & - \\ \cline{3-9}
 &  & A* & 0 & 0 & 2.5 & 0 & - & - \\ \cline{2-9} 
 & \multirow{4}{*}{3-hop} & H & 0.48 & 0 & 5.31 & 0 & 7.73 & 0 \\ \cline{3-9} 
 &  & H* & 0 & 0 & 2.9 & 0 & 8.7 & 0 \\ \cline{3-9}
 &  & A & 0 & 0 & 3.86 & 1.93 & 7.25 & 0 \\ \cline{3-9}
 &  & A* & 0 & 0 & 2.9 & 0.97 & 7.73 & 0 \\ \hline
\multicolumn{2}{|c|}{\multirow{4}{*}{PQL3}} & H & 0 & 0 & 3.86 & 0 & 7.73 & 0 \\ \cline{3-9} 
\multicolumn{2}{|c|}{} & H* & 0 & 0 & 2.9 & 0 & 7.73 & 0 \\ \cline{3-9} 
\multicolumn{2}{|c|}{} & A & 0 & 0 & 2.9 & 0.97 & 7.25 & 0 \\ \cline{3-9}
\multicolumn{2}{|c|}{} & A* & 0 & 0 & 2.9 & 0 & 7.73 & 0 \\ \hline

\multicolumn{2}{|c|}{\multirow{2}{*}{Train3, Test2}} & H & 1.13 & 2.95 & 2.7 & 53.58 & - & - \\ \cline{3-9} 
\multicolumn{2}{|c|}{} & A & 0.25 & 2.38 & 7.4 & 40.03 & - & - \\ \hline
\end{tabular}
}
	 \caption{Distribution of error types under \system
	 framework (in percentage). H stands for HR-BiLSTM, A for ABWIM, RE for
	 `Relation Extraction', and TD for `Termination Decision'.
	 * denotes the +DQ setting.}
    \label{tab:err3}
        \vspace{-1em}
    
\end{table}
The interpretability of \system, i.e., the possibility to analyze each hop, facilitates the analysis of error distributions.
We list the percentage of questions for which \system fails to extract the correct
relations by the number of hops for different datasets.
The results of HR\_BiLSTM and ABWIM within the \system framework are reported in
Table~\ref{tab:err3}.
Our observations are offered below.

First, whether for 1-hop or 2-hop WQ questions, both models suffer
in relation extraction in the first hop, whereas there are fewer errors in the second hop and for the termination decision.

Second, for the PQ/PQL datasets, as with the WQ dataset,
incorrect relation extraction is the major error, and surprisingly there were no
errors for termination decision except for a few on PQL3 with ABWIM.
After comparing the 2-hop testing data from PQ2/PQL2 and PQ+/PQL+, we also observe that long questions help the learning of short questions.
The model predicts better on 2-hop data when trained on both 2-hop and 3-hop
data than when trained on 2-hop data only. Here the improvement in relation
extraction in the first hop is the main contributor to this improved performance. 
In contrast, the performance on 3-hop data suffers when trained on 2-hop data.

Third, dynamic question representations (noted by *) significantly benefit the
relation extraction (RE) for the first hop. As \system utilizes the same model
for relation selection and termination decision, relieving the attention to the
previous relation in the later selection process in the training phase
decreases the ambiguity in the earlier selection process in the testing phase. 

Finally, in the experiments trained on 3-hop and tested on 2-hop, the model does not
terminate correctly on more than 40\% of the PQL2 data even though the relation
extraction for 1-hop and 2-hop are both correct. 
We conclude that having no samples of the predicted length for training still
hurts performance.
In addition, there are also a few early terminations after the first relation
extraction. Due to the different generation processes with different templates
for the 2-hop and 3-hop questions in PQL, learning from one may not apply to the
other.

\section{Conclusion}

In this paper, we propose the \system framework to allow an unrestricted number of
hops in knowledge-based relation extraction and to reduce the search space.
Results show that running the same model in the \system framework achieves
comparable results in a reduced search space. Moreover, experiments show \system
works well for lengthy relation extraction and can be applied to small, simple
KBs with task-specific relations. \system even facilitates the use of most
state-of-the-art models, and its transition-based design naturally supports the
dynamic question representation for better performance. 
These results attest its strong power for knowledge-based relation
extraction. The current framework uses a greedy search for each single hop.
We expect in the future that incorporating a beam search may further improve
performance.

\section*{Acknowledgement}
This research is partially supported by Ministry of Science and Technology, Taiwan under Grant no. MOST108-2634-F-002-008-, and the Academia Sinica Thematic Project under Grant no. 233b-1070100.

\bibliography{naacl2019_UHop}

\begin{thebibliography}{39}
\expandafter\ifx\csname natexlab\endcsname\relax\def\natexlab#1{#1}\fi

\bibitem[{Bast and Haussmann(2015)}]{bast2015more}
Hannah Bast and Elmar Haussmann. 2015.
\newblock More accurate question answering on {F}reebase.
\newblock In \emph{Proceedings of the 24th ACM International on Conference on
  Information and Knowledge Management}, pages 1431--1440. ACM.

\bibitem[{Berant et~al.(2013)Berant, Chou, Frostig, and
  Liang}]{berant2013semantic}
Jonathan Berant, Andrew Chou, Roy Frostig, and Percy Liang. 2013.
\newblock Semantic parsing on {F}reebase from question-answer pairs.
\newblock In \emph{Proceedings of the 2013 Conference on Empirical Methods in
  Natural Language Processing}, pages 1533--1544.

\bibitem[{Bordes et~al.(2015)Bordes, Usunier, Chopra, and
  Weston}]{bordes2015large}
Antoine Bordes, Nicolas Usunier, Sumit Chopra, and Jason Weston. 2015.
\newblock Large-scale simple question answering with memory networks.
\newblock \emph{arXiv preprint arXiv:1506.02075}.

\bibitem[{Chen et~al.(2018{\natexlab{a}})Chen, Chen, Huang, Ku, Chiu, and
  Yang}]{ChenRelation}
Hung-Chen Chen, Zi-Yuan Chen, Sin-Yi Huang, Lun-Wei Ku, Yu-Shian Chiu, and
  Wei-Jen Yang. 2018{\natexlab{a}}.
\newblock Relation extraction in knowledge base question answering: {F}rom
  general-domain to the catering industry.
\newblock In \emph{HCI in Business, Government, and Organizations}, pages
  26--41, Cham. Springer International Publishing.

\bibitem[{Chen et~al.(2018{\natexlab{b}})Chen, Xiong, Yan, and
  Wang}]{chen2018variational}
Wenhu Chen, Wenhan Xiong, Xifeng Yan, and William~Yang Wang.
  2018{\natexlab{b}}.
\newblock Variational knowledge graph reasoning.
\newblock In \emph{Proceedings of the 2018 Conference of the North American
  Chapter of the Association for Computational Linguistics: Human Language
  Technologies, Volume 1 (Long Papers)}, volume~1, pages 1823--1832.

\bibitem[{Das et~al.(2017)Das, Dhuliawala, Zaheer, Vilnis, Durugkar,
  Krishnamurthy, Smola, and McCallum}]{das2017go}
Rajarshi Das, Shehzaad Dhuliawala, Manzil Zaheer, Luke Vilnis, Ishan Durugkar,
  Akshay Krishnamurthy, Alex Smola, and Andrew McCallum. 2017.
\newblock Go for a walk and arrive at the answer: Reasoning over paths in
  knowledge bases using reinforcement learning.
\newblock \emph{arXiv preprint arXiv:1711.05851}.

\bibitem[{Dong et~al.(2015)Dong, Wei, Zhou, and Xu}]{dong2015question}
Li~Dong, Furu Wei, Ming Zhou, and Ke~Xu. 2015.
\newblock Question answering over {F}reebase with multi-column convolutional
  neural networks.
\newblock In \emph{Proceedings of the 53rd Annual Meeting of the Association
  for Computational Linguistics and the 7th International Joint Conference on
  Natural Language Processing (Volume 1: Long Papers)}, volume~1, pages
  260--269.

\bibitem[{Gardner et~al.(2014)Gardner, Talukdar, Krishnamurthy, and
  Mitchell}]{gardner2014incorporating}
Matt Gardner, Partha Talukdar, Jayant Krishnamurthy, and Tom Mitchell. 2014.
\newblock Incorporating vector space similarity in random walk inference over
  knowledge bases.
\newblock In \emph{Proceedings of the 2014 Conference on Empirical Methods in
  Natural Language Processing (EMNLP)}, pages 397--406.

\bibitem[{Gardner et~al.(2013)Gardner, Talukdar, Kisiel, and
  Mitchell}]{gardner2013improving}
Matt Gardner, Partha~Pratim Talukdar, Bryan Kisiel, and Tom Mitchell. 2013.
\newblock Improving learning and inference in a large knowledge-base using
  latent syntactic cues.
\newblock In \emph{Proceedings of the 2013 Conference on Empirical Methods in
  Natural Language Processing}, pages 833--838.

\bibitem[{Hao et~al.(2017)Hao, Zhang, Liu, He, Liu, Wu, and Zhao}]{hao2017end}
Yanchao Hao, Yuanzhe Zhang, Kang Liu, Shizhu He, Zhanyi Liu, Hua Wu, and Jun
  Zhao. 2017.
\newblock An end-to-end model for question answering over knowledge base with
  cross-attention combining global knowledge.
\newblock In \emph{Proceedings of the 55th Annual Meeting of the Association
  for Computational Linguistics (Volume 1: Long Papers)}, volume~1, pages
  221--231.

\bibitem[{Hinton et~al.(2014)Hinton, Srivastava, and
  Swersky}]{hinton2012neural}
Geoffrey Hinton, Nitish Srivastava, and Kevin Swersky. 2014.
\newblock {Notes from Lecture 6a of Neural Networks for Machine Learning:
  Overview of mini-batch gradient descent}.

\bibitem[{Iyyer et~al.(2017)Iyyer, Yih, and Chang}]{iyyer2017search}
Mohit Iyyer, Wen-tau Yih, and Ming-Wei Chang. 2017.
\newblock Search-based neural structured learning for sequential question
  answering.
\newblock In \emph{Proceedings of the 55th Annual Meeting of the Association
  for Computational Linguistics (Volume 1: Long Papers)}, volume~1, pages
  1821--1831.

\bibitem[{Kamigaito et~al.(2017)Kamigaito, Hayashi, Hirao, Takamura, Okumura,
  and Nagata}]{kamigaito2017supervised}
Hidetaka Kamigaito, Katsuhiko Hayashi, Tsutomu Hirao, Hiroya Takamura, Manabu
  Okumura, and Masaaki Nagata. 2017.
\newblock Supervised attention for sequence-to-sequence constituency parsing.
\newblock In \emph{Proceedings of the Eighth International Joint Conference on
  Natural Language Processing (Volume 2: Short Papers)}, volume~2, pages 7--12.

\bibitem[{Krishnamurthy et~al.(2017)Krishnamurthy, Dasigi, and
  Gardner}]{krishnamurthy2017neural}
Jayant Krishnamurthy, Pradeep Dasigi, and Matt Gardner. 2017.
\newblock Neural semantic parsing with type constraints for semi-structured
  tables.
\newblock In \emph{Proceedings of the 2017 Conference on Empirical Methods in
  Natural Language Processing}, pages 1516--1526.

\bibitem[{Lao et~al.(2011)Lao, Mitchell, and Cohen}]{lao2011random}
Ni~Lao, Tom Mitchell, and William~W Cohen. 2011.
\newblock Random walk inference and learning in a large scale knowledge base.
\newblock In \emph{Proceedings of the Conference on Empirical Methods in
  Natural Language Processing}, pages 529--539. Association for Computational
  Linguistics.

\bibitem[{Lin et~al.(2018)Lin, Socher, and Xiong}]{LinRX2018:MultiHopKG}
Xi~Victoria Lin, Richard Socher, and Caiming Xiong. 2018.
\newblock Multi-hop knowledge graph reasoning with reward shaping.
\newblock In \emph{Proceedings of the 2018 Conference on Empirical Methods in
  Natural Language Processing, {EMNLP} 2018, Brussels, Belgium, October
  31-November 4, 2018}.

\bibitem[{Liu et~al.(2016)Liu, Utiyama, Finch, and Sumita}]{liu2016neural}
Lemao Liu, Masao Utiyama, Andrew Finch, and Eiichiro Sumita. 2016.
\newblock Neural machine translation with supervised attention.
\newblock In \emph{Proceedings of COLING 2016, the 26th International
  Conference on Computational Linguistics: Technical Papers}, pages 3093--3102.

\bibitem[{Lukovnikov et~al.(2017)Lukovnikov, Fischer, Lehmann, and
  Auer}]{lukovnikov2017neural}
Denis Lukovnikov, Asja Fischer, Jens Lehmann, and S{\"o}ren Auer. 2017.
\newblock Neural network-based question answering over knowledge graphs on word
  and character level.
\newblock In \emph{Proceedings of the 26th international conference on World
  Wide Web}, pages 1211--1220. International World Wide Web Conferences
  Steering Committee.

\bibitem[{Mi et~al.(2016)Mi, Wang, and Ittycheriah}]{mi2016supervised}
Haitao Mi, Zhiguo Wang, and Abe Ittycheriah. 2016.
\newblock Supervised attentions for neural machine translation.
\newblock In \emph{Proceedings of the 2016 Conference on Empirical Methods in
  Natural Language Processing}, pages 2283--2288.

\bibitem[{Peng et~al.(2017)Peng, Chang, and Yih}]{peng2017maximum}
Haoruo Peng, Ming-Wei Chang, and Wen-tau Yih. 2017.
\newblock Maximum margin reward networks for learning from explicit and
  implicit supervision.
\newblock In \emph{Proceedings of the 2017 Conference on Empirical Methods in
  Natural Language Processing}, pages 2368--2378.

\bibitem[{Pennington et~al.(2014)Pennington, Socher, and
  Manning}]{pennington2014glove}
Jeffrey Pennington, Richard Socher, and Christopher~D. Manning. 2014.
\newblock \href {http://www.aclweb.org/anthology/D14-1162} {{GloVe: Global
  Vectors for Word Representation}}.
\newblock In \emph{Empirical Methods in Natural Language Processing (EMNLP)},
  pages 1532--1543.

\bibitem[{Saha et~al.(2018)Saha, Pahuja, Khapra, Sankaranarayanan, and
  Chandar}]{saha2018complex}
Amrita Saha, Vardaan Pahuja, Mitesh~M. Khapra, Karthik Sankaranarayanan, and
  Sarath Chandar. 2018.
\newblock \href
  {https://www.aaai.org/ocs/index.php/AAAI/AAAI18/paper/view/17181} {Complex
  sequential question answering: {T}owards learning to converse over linked
  question answer pairs with a knowledge graph}.
\newblock In \emph{Proceedings of the Thirty-Second {AAAI} Conference on
  Artificial Intelligence, New Orleans, Louisiana, USA, February 2-7, 2018}.

\bibitem[{See et~al.(2017)See, Liu, and Manning}]{see2017get}
Abigail See, Peter~J Liu, and Christopher~D Manning. 2017.
\newblock Get to the point: {S}ummarization with pointer-generator networks.
\newblock \emph{arXiv preprint arXiv:1704.04368}.

\bibitem[{Shen et~al.(2018)Shen, Huang, Liang, Li, and Fu}]{shen2018chinese}
Cun Shen, Tinglei Huang, Xiao Liang, Feng Li, and Kun Fu. 2018.
\newblock Chinese knowledge base question answering by attention-based
  multi-granularity model.
\newblock \emph{Information}, 9(4):98.

\bibitem[{Sorokin and Gurevych(2018)}]{socokin2018modeling}
Daniil Sorokin and Iryna Gurevych. 2018.
\newblock \href {http://aclweb.org/anthology/C18-1280} {Modeling semantics with
  gated graph neural networks for knowledge base question answering}.
\newblock In \emph{Proceedings of the 27th International Conference on
  Computational Linguistics}, pages 3306--3317. Association for Computational
  Linguistics.

\bibitem[{Toutanova et~al.(2015)Toutanova, Chen, Pantel, Poon, Choudhury, and
  Gamon}]{toutanova2015representing}
Kristina Toutanova, Danqi Chen, Patrick Pantel, Hoifung Poon, Pallavi
  Choudhury, and Michael Gamon. 2015.
\newblock Representing text for joint embedding of text and knowledge bases.
\newblock In \emph{Proceedings of the 2015 Conference on Empirical Methods in
  Natural Language Processing}, pages 1499--1509.

\bibitem[{Williams(1992)}]{williams1992simple}
Ronald~J Williams. 1992.
\newblock Simple statistical gradient-following algorithms for connectionist
  reinforcement learning.
\newblock \emph{Machine learning}, 8(3-4):229--256.

\bibitem[{Xiong et~al.(2017)Xiong, Hoang, and Wang}]{xiong2017deeppath}
Wenhan Xiong, Thien Hoang, and William~Yang Wang. 2017.
\newblock {DeepPath: A reinforcement learning method for knowledge graph
  reasoning}.
\newblock \emph{arXiv preprint arXiv:1707.06690}.

\bibitem[{Xu et~al.(2016)Xu, Reddy, Feng, Huang, and Zhao}]{xu2016question}
Kun Xu, Siva Reddy, Yansong Feng, Songfang Huang, and Dongyan Zhao. 2016.
\newblock Question answering on {F}reebase via relation extraction and textual
  evidence.
\newblock In \emph{Proceedings of the 54th Annual Meeting of the Association
  for Computational Linguistics (Volume 1: Long Papers)}, volume~1, pages
  2326--2336.

\bibitem[{Yang et~al.(2017)Yang, Yang, and Cohen}]{yang2017differentiable}
Fan Yang, Zhilin Yang, and William~W Cohen. 2017.
\newblock Differentiable learning of logical rules for knowledge base
  reasoning.
\newblock In \emph{Advances in Neural Information Processing Systems}, pages
  2319--2328.

\bibitem[{Yang and Chang(2015)}]{yang2015s}
Yi~Yang and Ming-Wei Chang. 2015.
\newblock {S-MART}: {N}ovel tree-based structured learning algorithms applied
  to tweet entity linking.
\newblock In \emph{Proceedings of the 53rd Annual Meeting of the Association
  for Computational Linguistics and the 7th International Joint Conference on
  Natural Language Processing (Volume 1: Long Papers)}, volume~1, pages
  504--513.

\bibitem[{Yih et~al.(2015)Yih, Chang, He, and Gao}]{yih2015semantic}
Wen-tau Yih, Ming-Wei Chang, Xiaodong He, and Jianfeng Gao. 2015.
\newblock Semantic parsing via staged query graph generation: {Q}uestion
  answering with knowledge base.
\newblock In \emph{Proceedings of the 53rd Annual Meeting of the Association
  for Computational Linguistics and the 7th International Joint Conference on
  Natural Language Processing (Volume 1: Long Papers)}, volume~1, pages
  1321--1331.

\bibitem[{Yih et~al.(2016)Yih, Richardson, Meek, Chang, and Suh}]{yih2016value}
Wen-tau Yih, Matthew Richardson, Chris Meek, Ming-Wei Chang, and Jina Suh.
  2016.
\newblock The value of semantic parse labeling for knowledge base question
  answering.
\newblock In \emph{Proceedings of the 54th Annual Meeting of the Association
  for Computational Linguistics (Volume 2: Short Papers)}, volume~2, pages
  201--206.

\bibitem[{Yin et~al.(2016)Yin, Yu, Xiang, Zhou, and
  Sch{\"u}tze}]{yin2016simple}
Wenpeng Yin, Mo~Yu, Bing Xiang, Bowen Zhou, and Hinrich Sch{\"u}tze. 2016.
\newblock Simple question answering by attentive convolutional neural network.
\newblock In \emph{Proceedings of COLING 2016, the 26th International
  Conference on Computational Linguistics: Technical Papers}, pages 1746--1756.

\bibitem[{Yu et~al.(2017)Yu, Yin, Hasan, dos Santos, Xiang, and
  Zhou}]{Moimproved}
Mo~Yu, Wenpeng Yin, Kazi~Saidul Hasan, C{\'{\i}}cero~Nogueira dos Santos, Bing
  Xiang, and Bowen Zhou. 2017.
\newblock \href {https://doi.org/10.18653/v1/P17-1053} {Improved neural
  relation detection for knowledge base question answering}.
\newblock In \emph{Proceedings of the 55th Annual Meeting of the Association
  for Computational Linguistics, {ACL} 2017, Vancouver, Canada, July 30 -
  August 4, Volume 1: Long Papers}, pages 571--581.

\bibitem[{Yu et~al.(2018)Yu, Hasan, Yu, Zhang, and Wang}]{Yu2018KnowledgeBR}
Yang Yu, Kazi~Saidul Hasan, Mo~Yu, Wei Zhang, and Zhiguo Wang. 2018.
\newblock Knowledge base relation detection via multi-view matching.
\newblock \emph{CoRR}, abs/1803.00612.

\bibitem[{Zhang et~al.(2018{\natexlab{a}})Zhang, Xu, Liang, Huang, and
  Fu}]{Zhang2018AnAW}
Hongzhi Zhang, Guandong Xu, Xiao Liang, Tinglei Huang, and Kun Fu.
  2018{\natexlab{a}}.
\newblock An attention-based word-level interaction model: {R}elation detection
  for knowledge base question answering.
\newblock \emph{CoRR}, abs/1801.09893.

\bibitem[{Zhang et~al.(2018{\natexlab{b}})Zhang, Dai, Kozareva, Smola, and
  Song}]{zhang2018variational}
Yuyu Zhang, Hanjun Dai, Zornitsa Kozareva, Alexander~J. Smola, and Le~Song.
  2018{\natexlab{b}}.
\newblock \href
  {https://www.aaai.org/ocs/index.php/AAAI/AAAI18/paper/view/16983}
  {Variational reasoning for question answering with knowledge graph}.
\newblock In \emph{Proceedings of the Thirty-Second {AAAI} Conference on
  Artificial Intelligence, New Orleans, Louisiana, USA, February 2-7, 2018}.

\bibitem[{Zhou et~al.(2018)Zhou, Huang, and Zhu}]{zhou2018interpretable}
Mantong Zhou, Minlie Huang, and Xiaoyan Zhu. 2018.
\newblock \href {https://aclanthology.info/papers/C18-1171/c18-1171} {An
  interpretable reasoning network for multi-relation question answering}.
\newblock In \emph{Proceedings of the 27th International Conference on
  Computational Linguistics, {COLING} 2018, Santa Fe, New Mexico, USA, August
  20-26, 2018}, pages 2010--2022.

\end{thebibliography}
\bibliographystyle{acl_natbib}
\end{CJK*}
\end{document}